\documentclass[letterpaper]{article} 
\usepackage{aaai25}  
\usepackage{xcolor}
\usepackage{graphicx}
\usepackage{caption}
\usepackage{subcaption}
\usepackage{times}  
\usepackage{helvet}  
\usepackage{courier}  
\usepackage[hyphens]{url}  
\usepackage{graphicx} 
\def\UrlFont{\rm}  
\usepackage{natbib}  
\usepackage{caption} 
\usepackage{algorithm}
\usepackage{algorithmic}

\newcommand{\revise}[1]{{\color{black} #1}}

\usepackage{newfloat}
\usepackage{listings}
\DeclareCaptionStyle{ruled}{labelfont=normalfont,labelsep=colon,strut=off} 
\floatstyle{ruled}
\newfloat{listing}{tb}{lst}{}
\floatname{listing}{Listing}
\title{Aligning AI with Public Values: Deliberation and Decision-Making for Governing Multimodal LLMs in Political Video Analysis}
\author{
Tanusree Sharma\textsuperscript{\rm 1},
    Yujin Potter\textsuperscript{\rm 3},
    Zachary Kilhoffer\textsuperscript{\rm 2},
Yun Huang\textsuperscript{\rm 2},
  Dawn Song\textsuperscript{\rm 3},
    Yang Wang \textsuperscript{\rm 2}
}
\affiliations{
    \textsuperscript{\rm 1} Pennsylvania State University\\
    \textsuperscript{\rm 2} University of Illinois at Urbana Champaign\\
     \textsuperscript{\rm 3} University of California, Berkeley\\

 \textsuperscript{\rm 1}   tanusree.sharma@psu.edu
}

\usepackage{bibentry}

\begin{document}

\maketitle

\begin{abstract}
How AI models should deal with political topics has been discussed, but it remains challenging and requires better governance. This paper examines the governance of large language models through individual and collective deliberation, focusing on politically sensitive videos. We conducted a two-step study: interviews with 10 journalists established a baseline understanding of expert video interpretation;  114 individuals through deliberation using \textit{Inclusive.AI}, a platform that facilitates democratic decision-making through decentralized autonomous organization (DAO) mechanisms. 
Our findings reveal distinct differences in interpretative priorities: 
 while experts emphasized emotion and narrative, general public prioritized factual clarity, objectivity, and emotional neutrality. Furthermore, we examined how different governance mechanisms - quadratic vs. weighted voting and equal vs. 20/80 voting power - shape users' decision-making regarding AI behavior. Results indicate that voting methods significantly influence outcomes, with quadratic voting reinforcing perceptions of liberal democracy and political equality. 
 Our study underscores the necessity of selecting appropriate governance mechanisms to better capture user perspectives and suggests decentralized AI governance as a potential way to facilitate broader public engagement in AI development, ensuring that varied perspectives meaningfully inform design decisions.
\end{abstract}

\section{Introduction}

A major criticism of AI development is the lack of transparency, particularly the insufficient documentation, and traceability in model design, specification, and deployment~\cite{brundage2020toward}, leading to adverse outcomes including discrimination, lack of representation, and breaches of legal regulations. Traditional social science approaches, such as interviews and surveys, often fall short in capturing user expectations due to their limitations in facilitating ongoing deliberation.
Governance, in contrast, is an interdisciplinary research area that involves stakeholders, 
~\cite{shneiderman2020bridging, bu2020privacy, rubinstein2013privacy, wang2022developing} for structural changes, such as defining bias criteria, determining rules for dataset diversity, etc. This involves principles such as normative positions, concrete actions, and engineering practices.

AI governance literature often clusters into key themes, many borrowed from data protection and privacy fields- (1) FACT - fairness, accuracy, confidentiality, and transparency~\cite{kemper2019transparent, kaminski2020multi, selbst2021institutional}; (2) FATE - fairness, accountability, transparency, and ethics~\cite{barocas2013governing}; (3) privacy preservation; (4) governance, compliance, and risk~\cite{calo2017artificial, gasser2017layered, scherer2015regulating, butcher2019state}; (5) trust and safety~\cite{biden2023executive, shneiderman113thNOTEHumanCentered2023, wang2022developing, saravanakumar2014survey, biden2023executive}; and (6) alignment with human values~\cite{ji2024beavertails, norhashim2024measuring}. Additionally, there is a growing focus on participatory AI~\cite{young2024participation} 
leveraging existing international legal frameworks~\cite{cihon2019standards, maas2021aligning, wallach2018agile, erdelyi2018regulating}. The AI Executive Order further highlights the need for a coordinated approach, emphasizing community engagement~\cite{biden2023executive}.

Emerging models such as Decentralized Autonomous Organizations (DAOs)~\cite{sharma2023unpacking} 
also provide innovative directions for technical elements that support varied structural concepts from management science and community coordination. DAOs are blockchain-based organizations governed by smart contracts and decentralized decision-making, enabling collective governance without centralized control~\cite{sharma2023unpacking}. By leveraging transparent, automated processes with smart contract governance, DAO provides a potential empirical testbed for exploring social choice experiments in potentially improving the current AI governance structure through a computational lens~\cite{benkler2015peer, lalley2018quadratic, weyl2022decentralized, zhang2017brief, weber2015realizing}. However, a fundamental tension exists between participatory decision-making in AI and its global, distributed nature~\cite{young2024participation}. DAOs present unique opportunities to address this challenge by implementing mechanisms such as social choice designs, quadratic voting, and liquid democracy~\cite{lalley2018quadratic, weyl2022decentralized, zhang2017brief}, while also enabling anonymous participation for diverse voices.

To examine the benefits of decentralized governance in AI development, we conduct a case study focusing on how AI systems should address politically sensitive topics. The use of LLMs in political domains has been widely debated, including their political biases~\cite{potter2024democracy,potter2024hidden,rozado2024political,feng2023pretraining,santurkar2023whose}. Recent studies have revealed that LLMs can influence users' political views through their interactions~\cite{potter2024hidden,fisher2024biased,costello2024durably}. While several approaches have been proposed to pursue the political neutrality of LLMs, no clear consensus has emerged~\cite{potter2024democracy,sorensenposition}; for example, many users expressed enjoyment when they are engaged in the interaction with politically leaned LLMs~\cite{potter2024hidden}. The conflicting views on these issues highlight the need for a deliberative process to incorporate diverse user perspectives.

This motivates our research questions: 

\textbf{RQ1: How does the general public perceive the use of LLM in political content interpretation? }

\textbf{RQ2: How do DAO governance mechanisms influence public opinions about improving LLM design?}

\begin{figure*}
    \centering
    \includegraphics[width=1.0\linewidth]{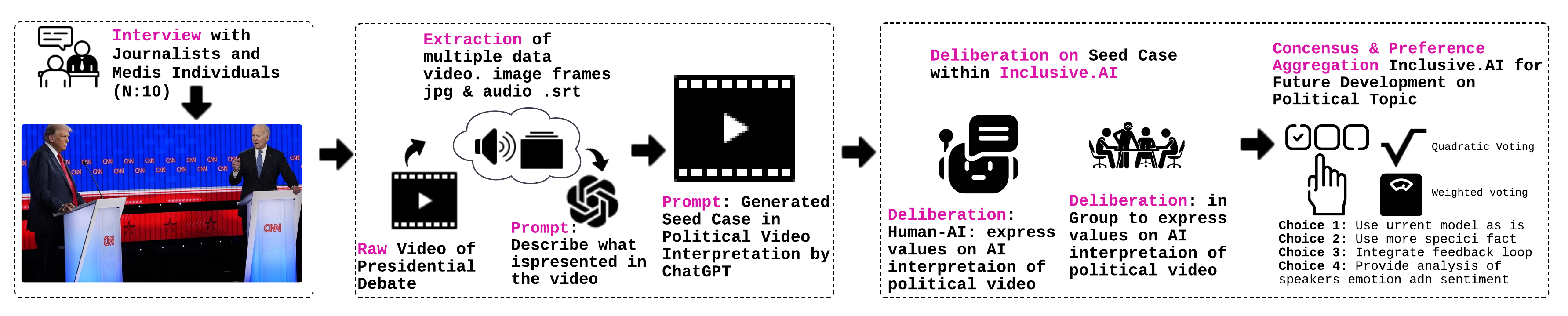}
    \caption{An overview of processes, (a) interview with experts to select suitable video example; (b) prepare seed case for experiment setup; (c) incorporate seed case into inclusive.AI system for deliberation and preference gathering (i) deliberation human-AI and group (ii) democratic voting process incorporating the voting options from deliberation of general public.}
    \label{fig:method}
\end{figure*}

We propose \textbf{Inclusive.AI}, a DAO-enabled governance, emphasizing inclusivity and human oversight in LLM design oversight. As illustrated in Figure~\ref{fig:method}, to explicitly understand users' specific expectations, the governance model allows users to deliberate on sensitive topics where LLM output can be controversial and contentious.
For our experiment, 
we used a video from the 2020 US presidential debate as a case study to explore public preferences in governing LLM behavior 
~\cite{linegar2023large}. To ensure secure and equitable participation, we implemented DAO infrastructure to enhance trust in the governance process. With Inclusive.AI, users first deliberate on LLM outputs, express their preferences and then participate in governance voting to guide future LLM design for political video interpretations.

\noindent\textbf{Findings.} Through an online experiment of 114 US internet users, our findings highlighted overlapping values between individual and collective deliberation for improving LLM output for political video content. Some factors are considered important, including, the emotions of the speaker, subjective content (e.g., who supports or opposes, composure, professionalism), and the speaker's positionality. There are some distinct differences in interpretative priorities: while experts emphasized emotion and narrative, general public prioritized factual clarity, objectivity, and emotional neutrality. 
Our findings also highlighted participants' perceived quality of the governance of the Inclusive.AI tool 
whereas voting methods significantly influence outcomes, with quadratic voting reinforcing perceptions of liberal democracy and political equality. They emphasized that quadratic voting, under equal voting power conditions, reduces the likelihood of producing unexpected outcomes compared to weighted voting. However, some were skeptical about whether the decided outcomes would be implemented in LLM models, suggesting guidelines at the government level to ensure compliance.

\section{Related Work}

\subsection{Video Analysis in Practice \& Multimodal Generative Vision Models.}
Videos are a rich source of information for communication~\cite{chen2019motion, lin2021vx2text}, driving tasks like video captioning, question answering~\cite{yang2021just, tseng2025biv, sharma2023disability}, text-video retrieval~\cite{gabeur2020multi, bain2021frozen, anne2017localizing}. 
Identifying key visual content in video-language learning remains a challenge~\cite{buch2022revisiting, lei2022revealing}. Political science research increasingly explores video content~\cite{hong2021analysis} where language models often exhibit biases in multi-modal data. 
Advancements in computer vision have led to foundational vision-language models, such as CLIP in numerous downstream applications, ranging from object detection to 3D applications~\cite{bangalath2022bridging, liang2023open, rozenberszki2022language, ni2022expanding}, and adapted for video applications~\cite{ni2022expanding, wang2021actionclip, rasheed2023fine}.
More recently, multimodal integration has advanced with models like Flamingo~\cite{alayrac2022flamingo}, BLIP-2~\cite{li2023blip} 
MiniGPT-4~\cite{zhu2023minigpt}, and LLaVA ~\cite{liu2024visual} leveraging web-scale image-text data for improved multimodal chat capabilities. Some works extend LLMs for video comprehension~\cite{maaz2023video, radford2021learning, chiang2023vicuna, li2023videochat, liu2024visual}, introducing Video-ChatGPT, model combining a video-optimizer for enhanced understanding.

\subsection{AI Governance Approaches}
 AI governance concerns putting values or principles into practice via policies, while values define what agents (people or AI) ought to do~\cite{shenBidirectionalHumanAIAlignment2024}. Researchers have argued that the research community broadly agrees about certain values and principles for better AI \cite{fjeldPrincipledArtificialIntelligence2020}, though the question of how best to implement these principles is far from solved, and proceeding at different paces, in fits and spurts, around the world.
 
\textbf{Regulatory Effort.} Historically, much of the focus in AI governance research has been at the national and subnational levels~\cite{calo2017artificial, gasser2017layered, scherer2015regulating}. 
US efforts in AI governance currently focus on executive actions and industry collaboration. 
Executive Order 14011 ``on the Safe, Secure, and Trustworthy Development and Use of Artificial Intelligence'' (AI Executive Order)~\cite{thewhitehouseExecutiveOrderSafe2023} highlights the dual nature of AI's potential, stating that while AI can address critical challenges and enhance prosperity, productivity, innovation, and security, it also entails risks that require careful regulation.
The U.S. approach to AI governance aims to ensure both safe and ethical AI development while achieving national strategic goals. In the short term, the U.S. will work with AI developers and other stakeholders to establish standards and guidelines. 

Executive Order 14011 -- like the more ambitious EU AI Act \cite{europeancommissionProposalREGULATIONEUROPEAN2021a} -- directs an assortment of actors to participate in the \emph{standardization} of AI development and deployment. Standards are formal documents that guide organizations in achieving specific goals such as interoperability, safety, or regulatory compliance. They provide a common language and practical guidelines for ensuring the safety and auditability of technical systems~\cite{wang2022developing}.
The most rigorous standards are intended to facilitate third-party audits and certification, signaling good and trustworthy practices~\cite{saravanakumar2014survey}.
At present, however, the two available AI standards from NIST (AI Risk Management Framework ~\cite{nistArtificialIntelligenceRisk2023}) and ISO (ISO 42001 AI Management System ~\cite{isoISOIEC42001}) are not auditable and not yet rigorously tested.


\textbf{Participatory approaches.} 
Driven by calls from civil society organizations, academia, and others, there is growing emphasis on participatory AI governance to make AI/ML design more inclusive and equitable~\cite{young2024participation, zhangDeliberatingAIImproving2023}.
However, there is limited empirical literature on involving stakeholders in refining AI performance. \revise{Frameworks like WeBuildAI~\cite{lee2019webuildai}, which facilitate participatory design for community-serving algorithms such as on demand food donation transportation services—illustrate the potential of inclusive approaches. Similarly, tools like ConsiderIt~\cite{fan2020digital}, which incorporate representative deliberation in content moderation decisions, underscore the value of incorporating user voices and interests in decision-making. Crowdsourcing, as explored by Lee et al.\cite{lee2014crowdsourcing}, demonstrates how participatory approaches can enhance democracy by optimizing the aggregation of individual preferences into collective decisions while fostering creativity\cite{salganik2015wiki}.}
Both in theory and practice, there exists a tension between the goal of participatory decision-making in AI and the global, distributed nature of AI development~\cite{young2024participation}.
Kemp et al.~\cite{erdelyi2018regulating} and some researchers advocate for decentralized approaches, such as \textit{``Governance Coordinating Committees,''} global standards, or leveraging existing international legal frameworks~\cite{cihon2019standards, maas2021aligning, wallach2018agile}.

Yet, current AI refinement and training processes, including reinforcement learning, often rely on labor pools from developing countries due to cost considerations~\cite{schmidt2019crowdsourced}. 
This dynamic presents a significant challenge, though not an insurmountable one, in ensuring meaningful participation on a suitable scale in AI development~\cite{young2024participation}.
In designing AI models, it is essential to involve stakeholders and affected communities, along with AI companies, to deliberate on sensitive AI-related topics and make informed decisions about AI model behavior. 
However, another difficulty is establishing the best role for AI to play in group decision-making~\cite{zhengCompetentRigidIdentifying2023}.

\subsection{DAO as a Tool for Governance and Co-ordination.}
Decentralized Autonomous Organizations (DAOs), which emerged in the mid-2010s, share commonalities with early online communities, especially those focused on open-source projects~\cite{chohan2017decentralized}. DAOs also draw inspiration from various models, including digital and platform cooperatives~\cite{mannan2018fostering}, multi-organizational networks like keiretsus~\cite{lincoln1996keiretsu}, crowdfunding platforms such as Patreon, virtual economies in games like World of Warcraft and Second Life~\cite{lehdonvirta2014virtual}, and peer-produced projects like Wikipedia~\cite{xu2015empirical}. DAO governance, as a human-centric digital organization, addresses key issues in social computing but can be more complex than platforms such as civic tech~\cite{poor2005mechanisms}, and traditional online communities~\cite{love2010linux}. DAOs were designed to automate organizational processes leveraging cryptographically secured blockchain technology~\cite{buterin2014daos}. A key function of a DAO is collective decision-making, carried out through a series of proposals where members vote on organizational events using governance tokens, signifying relative influence within the DAO. Voting mechanisms like weighted and quadratic voting ensure secure, pseudonymous participation, with voters identified by on-chain addresses rather than real-world identities.

The emergence of DAOs introduces possible solutions, including classic coordination dilemmas such as preference aggregation, credible commitments, audience costs, information asymmetry, representation, and accountability~\cite{hall1996political, openai}. The relevance of these theories to the design of digitally-native governance institutions is a critical question~\cite{rousseau1964social, dahl1989democracy, landemore2012democratic}. The separation of powers in DAOs helps prevent power concentration, enhance transparency, and mitigate organizational gridlock~\cite{de1989montesquieu}. This is increasingly relevant for AI, where inclusive decision-making is crucial throughout development lifecycle. In this work, we explore the design of DAO in AI governance for model decision-making.

\vspace{-2mm}
\vspace{-2mm}
\label{method}
\begin{figure*}[t!]
    \centering
    \includegraphics[width=\linewidth]{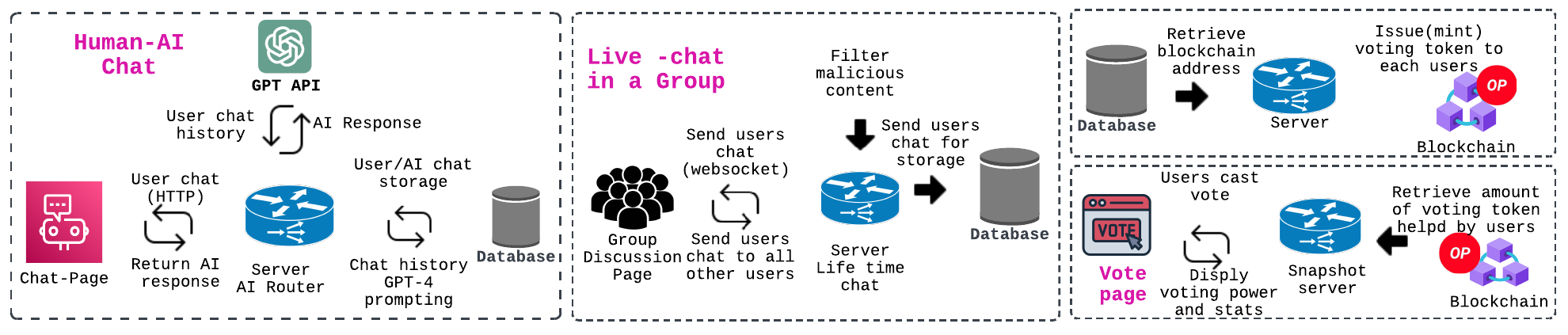}
    \caption{Incluisve.AI System Architecture
    } 
    \label{fig:system}
\end{figure*}


\section{Inclusive.AI Design and Experiment}

As shown in Figure~\ref{fig:method}, our entire study includes (1) an expert interview (protocol in Appendix~\ref{experts}) with journalists and media individuals in selecting a suitable political video~\footnote{Since we aim to understand the general public’s perception of the use of LLMs for sensitive topics, such as political content, selecting politically sensitive content for the study requires careful consideration. We leveraged experts’ opinions to conform to the inclusion criteria for selecting content) by providing them with an overview of the user study goal. The inclusion criteria mentioned: (a) relevance to current events (b) Broad political video (c) contextual depth or complexity  (d)  authenticity of content sources. We also asked them how they would prompt the LLM tool to interpret this video. We leveraged experts’ feedback to design the deliberation case.} as a seed case for user experiments; (2) a large-scale user experiment in deliberating users' values regarding the LLM interpretation of political topics 
The user experiment has three main design components-(1) Human-AI interaction to deliberate on sensitive topics (e.g. presidential debate video), (2) Group discussion to engage with other to understand collective opinions (3) Governance decisions to guide future LLM model updates. 
\subsection{System Design}
\label{system}
Inclusive.AI (GitHub~\cite{InclusiveAI}) democratic platform (Figure~\ref{fig:system}) is deployed on the Optimism blockchain and integrates with a custom server, using Web3Auth~\cite{goldreich1998secure} for authentication. Web3Auth generates a unique Multi-Party Computation (MPC) wallet for each user, derives their blockchain address, and enables message authentication for verifying participation in votes. Upon signup, they are guided to an introduction 2-minute video overview of task details and app functionality. They then proceed to Human-AI deliberation and group discussions, where a chat box with websocket connections supports real-time interactions.


For the voting page, we implemented two VoteToken contracts using Solidity, a programming language for the Ethereum blockchain to represent users' voting power. These tokens are \textit{minted} to users' accounts, allowing them to vote on proposals for LLM improvement of political video. The system uses the Snapshot API to create a space for governance and ensure all the processes are transparent in Blockchain. Spaces define voting rules (e.g., duration), proposal criteria (e.g., success thresholds of proposed options to be considered for LLM improvements), and roles for admins and moderators, including who can vote or propose changes.
We designed spaces for each experimental condition (each type of governance decision mechanism discussed in section~\ref{exp}). When the
user allocates votes accordingly and clicks the “Cast Vote” button (in Figure~\ref{fig:vote} in the Appendix), this triggers Web3Auth’s signing
library, which signs a message for Snapshot voting.


\subsection{Deliberation and Decision Making}
\label{design-deliberation}



\textbf{AI Guided Individual \& Group Deliberation.}
The app begins by engaging users with an AI Value Topic related to data interpretation of a video on a political topic by GPT4 (Figure~\ref{user-workflow} in the Appendix). This topic is based on a 6-minute clip from the 2020 US presidential debate~\cite{InclusiveAI}
The app presented a simple question: \textit{``Do you find the interpretation useful?'' with three options (yes, no, maybe)} to stimulate further thought. Based on the user's response to the provided options, the AI continues the corresponding chat that allows users to clarify their intentions and values in natural-language conversations about AI value topics. AI resolves ambiguities through multi-turn conversations, seeking clarifications and guiding users to define their norms and expectations. 
Following that, users engaged in a group deliberation and learned the perspectives of others' norms (Figure~\ref{fig:discussion} in the Appendix). 
This group deliberation enables users to co-validate their values with a mini-public to make informed decisions.
If participants are unable to introduce a topic on their own, they are encouraged to refer to the suggested topics provided by the tool. We designed the suggested topic based on the pilot experiment (in Appendix Section~\ref{pilot})

\noindent\textbf{Democratic Decision Making for Future MM-LLM}
Finally, users participate in a democratic decision-making process by voting.
We designed experiments to assess varying voting methods and combinations of voting power (details in section~\ref{exp}) to examine users' perception of the quality of the process being democratic in LLM model improvement decisions. 
We assessed users' self-reported quality with the Variety of Democracy (V-Dem) scale~\cite{lindberg2014v}. The voting was live for 48 hours. 

\subsection{User Experimental Design}
\label{exp}

\textbf{Treatment Condition: Varying Governance Voting Design} 
In governance decision-making, voting methods and voting power are key factors influencing outcomes, as demonstrated in DAOs and deliberative democracy~\cite{sharma2023unpacking, sharma2024future, fritsch2024analyzing, willis2022deliberative, follesdal2010place}. To structure decision-making to aggregate people's preferences for future LLM development, we designed a 2×2 treatment condition based on two factors: voting method and voting power, each with two levels. While alternative methods like single-choice or approval voting could also be considered, it would significantly increase the number of treatment conditions and require a large participant pool to achieve statistically significant results with actionable interpretations.

More specifically, we implemented weighted voting, commonly used in DAOs~\cite{sharma2023unpacking}, where users distribute voting power across multiple options based on preference. To counterbalance traditional democratic aggregation which may disadvantage minority views, we incorporated quadratic voting - largely applied in real-world cases, such as Gitcoin’s grant funding for public goods\cite{miller2024case}--which enhances minority influence on crucial issues by allowing users to ``pay'' for additional votes. For instance, with quadratic voting, 4 tokens provide 2 votes, emphasizing the number of voters rather than voting power size~\cite{lalley2016quadratic}. 
To address voting power distribution, we compared equal distribution with a Pareto-based 20/80 split, where 20\% of participants receive 80\% of tokens, simulating early adopters' influence. This model reflects real-world AI deployment scenarios, where certain groups benefit disproportionately.

Thus, there were four treatment conditions- (1) Quadratic Voting token-based (Participants having the same amount of token/voting power); (2) Quadratic Voting 20\% population get 80\% of the token as early adopters; (3) Weighted voting Token based (participants having the same amount of token/voting power); (4) Weighted voting 20\% population get 80\% of the token as early adopters. 
The goal was to assess how these variations influence users' perceptions of the process's democratic quality and outcome.


\noindent\textbf{Experimental Design.}
Participants were randomly assigned by the Inclusive.AI system to one of four governance decision-making mechanisms, forming four treatment groups. Participants didn't know the treatment group to which they had been assigned. We employed a $2*2$ between-subjects design with 114 participants (26-30 per condition). Participants voted on four MM-LLM update options derived from 20 pilot studies for political video interpretation: (i) keep the current model; (ii) provide more specific facts; (iii) integrate a user feedback loop; (iv) analyze speakers' emotions and sentiment (as shown in Figure~\ref{fig:vote}).

\begin{table}[h]
\centering
\caption{Experts demographics and background.}
\begin{tabular}{l l l l }
\hline
\textit{ID} & \textit{Gender} & \textit{Age} & \textit{Media Background} \\
\hline
E1 & Female & 25-34 & TV News, Police issues \\
E2 & Female & 25-34 & Environment, Architecture\\
E3 & Female & 25-34 & Local, Under-represented \\
E4 & Male & 35-44 & Political, Election \\
E5 & Female & 35-44 & Weather, Political \\
E6 & Male & 35-44 & TV Media\\
E7 & Female & 25-34 & Economy, Tesla\\
E8 & Male & 45-54 & Public Communication \\
E9 & Male & 25-34 & Political\\
E10 & Male & 25-34 & Local, Urban Design\\
\hline
\end{tabular}%
\label{table:demographics-ex}
\end{table}
 
 \begin{table*}[ht]
\centering

\caption{Participants' demographics ($n=114$) 
}
\label{tab:demographic}
\begin{tabular}{ccc cccc ccccc}

\multicolumn{3}{c}{\textbf{Gender (\%)}}&\multicolumn{4}{c}{\textbf{Age (\%)}}&\multicolumn{5}{c}{\textbf{Race (\%)}}\\
\hline
Woman &Man&Non-binary &18-24&25-34&35-44&45-54&White &Black & Asian & Latin & Others\\
45.6&52.6&1.8&21.1&39.5&27.2&12.3 & 52.6&12.3&21.9&10.5&2.63\\

\multicolumn{12}{c}{\textbf{Education (\%)}}\\
\hline
\multicolumn{1}{c}{High school}&\multicolumn{1}{c}{Bachelor}&\multicolumn{3}{c}{Masters/professional}&\multicolumn{2}{c}{Doctorate}&\multicolumn{4}{c}{College/vocational training}&\multicolumn{1}{c}{Others}\\
\multicolumn{1}{c}{14.0}&\multicolumn{1}{c}{41.2}&\multicolumn{3}{c}{12.3}&\multicolumn{2}{c}{2.6}&\multicolumn{4}{c}{26.3}&\multicolumn{1}{c}{3.5}\\

\multicolumn{12}{c}{\textbf{Political Orientation (\%)}}\\
\hline
\multicolumn{2}{c}{Very conservative}&\multicolumn{2}{c}{Conservative}&\multicolumn{3}{c}{Moderate}&\multicolumn{2}{c}{Liberal}&\multicolumn{3}{c}{Very liberal}\\
\multicolumn{2}{c}{5.3}&\multicolumn{2}{c}{21.1}&\multicolumn{3}{c}{22.8}&\multicolumn{2}{c}{35.1}&\multicolumn{3}{c}{15.8}\\

\multicolumn{12}{c}{\textbf{Political Party (\%)}}\\
\hline
\multicolumn{2}{c}{Republic party}&\multicolumn{2}{c}{Democratic party}&\multicolumn{3}{c}{Libertarian party}&\multicolumn{5}{c}{Independent/Unaffiliated}\\
\multicolumn{2}{c}{21.9}&\multicolumn{2}{c}{50.9}&\multicolumn{3}{c}{2.6}&\multicolumn{5}{c}{24.6}
\end{tabular}
\end{table*} 
\subsection{Participant Demographics}
We recruited participants who are USA residents. We recruited through the CloudResearch platform~\cite{CloudResearch}. This study protocol involving human subjects was approved by the Institutional Review Board (IRB). Each received \$30 for their participation. We used a set of screening questions. Respondents were invited to our study if they met all three selection criteria - (1) 18 years or older; (2) country of residence USA; (3) use generative AI tool. 
Our study resulted in total of 114 participants (Demographics in Table~\ref{tab:demographic}).

\subsection{Experts Background.}
text, images, and videos, particularly analyzing complex and sensitive topics, like US presidential debates. 
We recruited 10 US-based experts through personal connections and word of mouth. Experts in this study come from diverse backgrounds in journalism, media, and communication, with an equal split between males and females. Among them are
Ph.D researchers specializing in media studies in urban design, and political economy, journalist focused on video media who had experience in the 2020 election coverage; medical misinformation within local communities and journalists and videographers who have covered Tesla, police issues, and local TV media, offering a unique blend of skills and
perspectives


\subsection{Details: Experts' Interview}
\label{expert-interview}
 Our two primary objectives of the experts' interview (Table~\ref{table:demographics-ex}) were: (i) to have a baseline of how experts envision the use of MM-LLMs for interpreting political videos to the general public and (ii) to incorporate expert feedback into the development of our methodological approach, including criteria for selecting video examples for the study. Each interview took around 1 hour.

In the first set of questions, we inquired about their primary expertise and experience with various data types, including video. This helped us understand their approach to handling different media, covering real-time events, managing diverse data for tasks like media report writing, and the factors that influence the quality of their reporting. In the next set of questions, we showed them a political video of US presidential debate and asked them to interpret it using multimodal data (e.g., audio, visuals, closed captions). We asked, \textit{``Can you walk me through the process you employ to analyze the video content to write a report?''} Following this, we asked their opinion on using LLMs for video analysis 
We then showed them how LLMs (ChatGPT) interpreted the same video and asked for their thoughts to identify the benefits, limitations, and critical factors in interpreting contentious topics. 

Since we aim to understand the general public's perception of the use of MM-LLMs in better designing models for sensitive topics, such as political content, selecting politically sensitive content for the study requires careful consideration. We leveraged experts' opinions to conform to the inclusion criteria for selecting content (details in Appendix) by providing them with an overview of the user study goal.
We also asked them how they would prompt the LLM tool to interpret this video. We leveraged experts' feedback to design the deliberation seed usecase. 

\subsection{Data Analysis} 
%
For qualitative data, two researchers independently read through 20\% of the individual deliberation with AI-agent and group discussion data, developed codes,
and compared them until we developed a consistent codebook. We met regularly to discuss the coding and agreed on a shared codebook. Once the codebook was finalized, two researchers divided the remaining data and coded them. After completing coding for all individual and group deliberation, both researchers spot-checked other’s coded transcripts and did not find any inconsistencies. Finally, we organized our codes into higher-level categories. We followed a deductive thematic analysis to explore participants’ values, expectations towards MM-LLM video interpretation~\cite{clarke2017thematic}. We grouped lower-level codes into sub-themes and further extracted main themes. \revise{We used a similar approach for the qualitative analysis of expert interview data to identify high-level themes.}
For the quantitative analysis of users' perceptions of DAO governance mechanisms, the data analysis process is explained directly within the respective results.

 \begin{table*}[h!]
\centering
\caption{Overview of Themes of Deliberation on LLM Output of Political Video}
\label{table:scenario}
\begin{tabular}{p{3cm} p{12cm} p{1.5cm}}
\hline
\textbf{Theme} & \textbf{Quote} & \textbf{Ind / Group} \\
\hline
Emotion of Speakers & \textit{``There was a heated argument in video, both speakers didn’t want to give way for other to speak, Trump and moderator were talking like they were fighting, its not in the LLM output.''} & Indv \& Group\\
Objectivity of Situation & \textit{``It didn’t understand situation at all, AI was superficial, capturing the scene, distinguished between speaker and their political view is important, I could have just read the subtitle instead.''} & Indv \& Group\\
Desire FactChecking & \textit{``Fact-checking whether the debaters are saying anything of substance would greatly help in giving an accurate picture of the view, like citing some source while interpreting the video.''} & Indv \& Group\\
Balance Brevity and Substance & ``AI describe he video, but there was no real context, like to take away'' & Group\\
Balancing Content \& Biases&  ``This is one of those times when I wish AI could let itself loose just a little more, necessarily—just to the fact of acknowledging how Trump was not acting as a good steward of discussion, also the too much emphasize towards Obamacare and social support system.''& Group\\ 
Organization of LLM output & ``It would be easier to differentiate to have the description of two candidates side by side.'' & Indv\\
Specific Design Recommendations, & ``Red highlight for content in the video that are factually wrong and green for truth.'' & Indv\\

\hline
\end{tabular}%
\label{tab:scenarios}
\end{table*}


\section{People's Opinion on LLM Interpretation of Political Video}
\subsection{Journalist's Opinion of LLM in Political Video.}
We found several practices of journalists in interpreting political videos on their own, including- (a) fact-checking with multiple data sources and guidelines (e.g. media literacy project~\cite{literacy}, MSA Security~\cite{msa}), (b) involvement of expert-in-the-loop (e.g. academic scholars, senior journalists, domain experts), (c) narrative approach considered as news generation 101; (d) theoretical underpinning, such as positionality, selective exposure~\cite{tully2022defining}.

Experts highlighted several limitations in LLM-generated summaries of political videos, particularly the absence of human interaction cues such as tone and emotion. They noted that the lack of contextual information, including background knowledge on political debates, reduced the summary’s usefulness for news content. While factually accurate, the summary failed to capture the antagonistic and dramatic dynamics of the debate, including conflicts, personal attacks, and the candidates' lack of factual references. Additionally, experts criticized its lack of storytelling and engagement, making it unsuitable for a diverse audience and insufficient in depth and impact.

\subsection{General Public's Opinion.} Our findings of users' interaction with the seed case on political video interpretation highlight various factors participants considered important on interpreting video content while analyzing multiple types of data (e.g. image frame, audio, etc). In group deliberation, we found that participants articulated their arguments in longer sentences, while in human-AI chats, the conversations were shorter. In individual value elicitation, we also found participants to suggest specific design recommendations of how to generate and present the LLM output rather than only pointing out what is lacking. They tend to begin their interactions with a positive tone. As the conversations progressed, participants shifted towards making recommendations and expressing concerns. In contrast, group deliberation started with a tone of concern and debate.

However, in both types of interactions, there are overlapping values emerged regarding LLM improvement for political content interpretation. 
This includes: the emotions of the speaker, subjective content (e.g., who supports or opposes, composure, professionalism), and the speaker's positionality. We also observed nuanced differences in individual values, for instance, participants tend to express a preference for fact-focused political LLM interpretation with specific indicators as design recommendations and emphasized the importance of clarity and organization of LLM output (Table~\ref{tab:scenarios} presents example quotes). 


\section{Experience in Democratic Governance}
\label{experience-chapter9}

\subsection{Preference on LLM Improvement Choices.}
For improving MM-LLMs in political video interpretation, participants strongly preferred \textit{``providing more specific facts''}(choice 2), followed by \textit{``analyzing speakers' emotions and sentiment''} (choice 4) and \textit{``integrating a user feedback loop''} (choice 3). The consistency of choices 2 and 4 across quadratic and weighted methods indicates stable user preference (Table~\ref{tab:condition-stat}). However, in 20/80 voting power distributions, early adopters (80\% power) influenced outcomes, narrowing the gap between choices 3 and 4. This suggests that in real-world governance of LLM improvements, decision-making that concentrates power among a few influential stakeholders could disproportionately shape LLM improvements, potentially misaligning with broader user preferences.

To see whether participants affiliated with different political parties had different choices and perceptions for LLM improvement on political video interpretation, we ran a linear regression controlling for voting methods. As a result, we found that, compared to Democrats, Republicans were less likely to vote for Choice 2 (i.e., provide more specific facts) with a P-value of $0.084$
and more likely to vote for Choice 3 (i.e., integrate a user feedback loop) with a P-value of $0.054$.

\begin{table*}[!htp]
\centering
\caption{Summary stats of the ratio of tokens allocated to each voting choice (Choice 1: Keep the current model, Choice 2: Provide more specific facts, Choice 3: Integrate a user feedback loop, and Choice 4: Analyze speakers’ emotions and sentiment) by users.
The ratio is calculated as the percentage of tokens the user allocated to each voting option. For example, if a user allocated 20, 20, 30, 30 tokens for each voting option, the vector for the user would be (0.2,0.2,0.3,0.3). 
} 
\label{tab:condition-stat}
\small
\begin{tabular}{rcccccccc}
&\multicolumn{2}{c}{\textbf{Choice 1}}&   \multicolumn{2}{c}{\textbf{Choice 2}} &\multicolumn{2}{c}{\textbf{Choice 3}}&   \multicolumn{2}{c}{\textbf{Choice 4}} \\
\hline
&mean &std.& mean &std&mean &std& mean &std. \\
\hline
Quadratic - same (n: 29)&0.0814 & 0.0885 & 0.4300 &0.3421& 0.1524 & 0.1690 & 0.2597 & 0.2905 \\
Quadratic -20/80 (n: 30)& 0.1193 & 0.2197 & 0.3267 & 0.2260 &  0.2188 & 0.1956 & 0.3080 & 0.2473\\
Ranked - same (n: 27) & 0.1193 & 0.1412 & 0.3941 & 0.2255 & 0.1896 & 0.1197 & 0.2926 & 0.2351\\
Ranked - 20/80 (n:28) & 0.1395 & 0.1534 & 0.4044 & 0.2692 & 0.1877 & 0.1400 & 0.2417 & 0.1870\\
\hline
\end{tabular}
\end{table*}

\begin{figure}[h]
    \centering
    \includegraphics[width=1.1\linewidth]{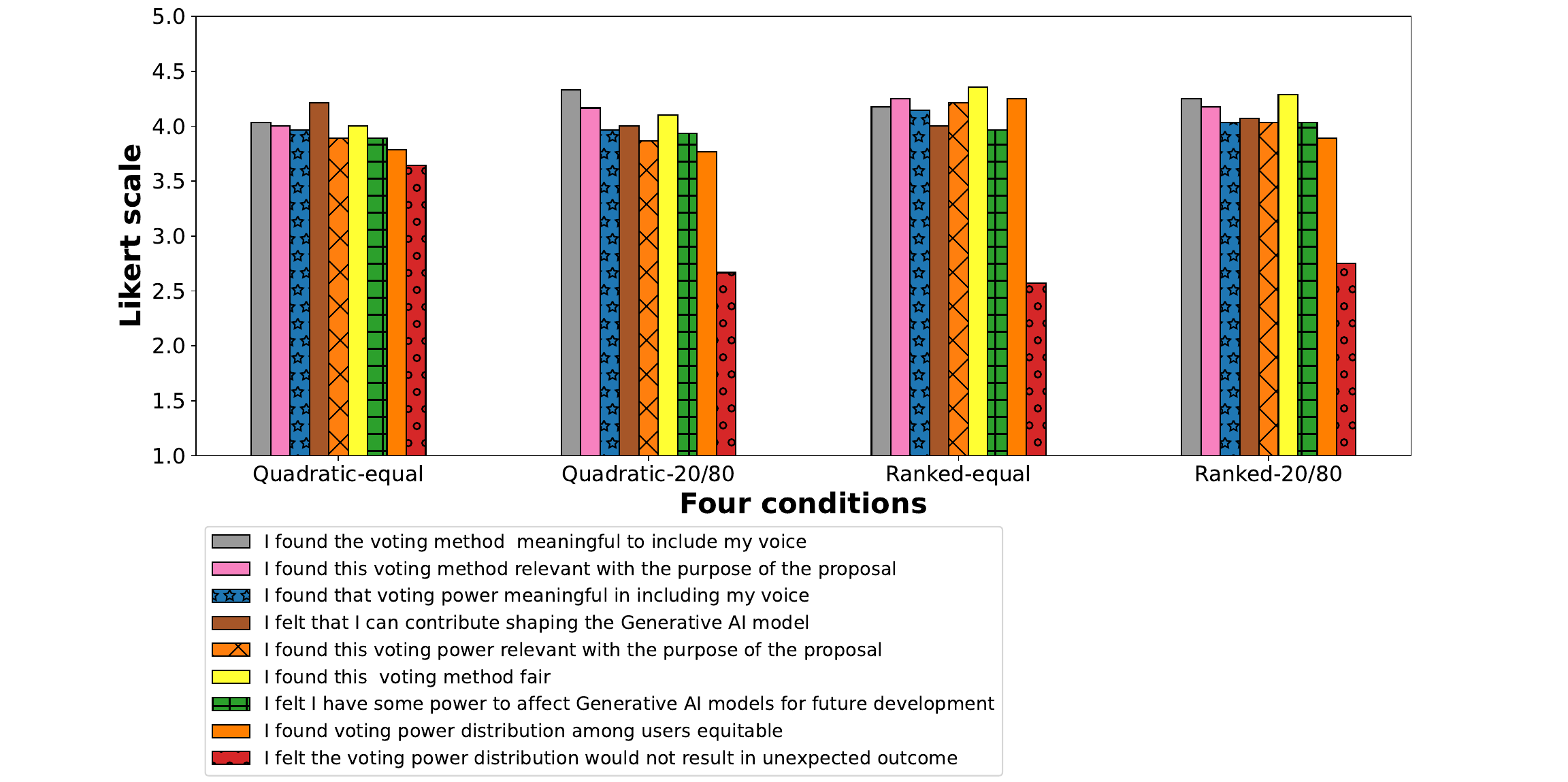}
    \caption{Users’ perception of the quality of voting mechanism in governance decision making}
    \label{fig:q2}
\end{figure}

\begin{table*}[]
\caption{Governance Decision-Making Experience Across Different Political Leaning}
\begin{tabular}{p{5cm} p{5cm} p{5cm}}

\textbf{Democrats}                                                                                 & \textbf{Republicans}                                                                            & \textbf{Independent / Unaffiliated}                                                         \\
\hline
Progressive and Empowerment                & Flexible in Distributing power                                                 & Desire for Additional Option   \\

Ease of Use and Intuitive                                                             & New Experience and  Curiosity &Ease of Use and Intuitive                                                         \\
Support Multiple Choices & Quantifying Perception and Thinking Critically & Weighted Voting as  a Preferred Feature\\

Quadratic Voting Perceived as Fair           & Having Influence  on AI Development        &  Applying This Process  to Other Contexts \\
 Engaging and Enjoyable                                                  &Concerns About Complexity and Restrictions  & Concerns of Fairness  and Transparency  \\
Informed and accurate  decision-making      &  Concerns About External Influences and Bias & Concerns About the Process’s Impact     \\
                                                                                                               \hline                     
\label{tab:politic}
\end{tabular}
\end{table*}
 
\subsection{General Perception of Voting Mechanism.}
With a 5-point Likert scale (Figure~\ref{fig:q2} in Appendix), we found participants' perceptions of the voting process usage in LLM governance where most participants were satisfied with the process regardless of voting protocols. Notably, they rated with average scores of $3.89, 4.17, 3.96$, and $3.93$ in the four voting mechanisms: quadratic$+$equal, quadratic$+$20/80, ranked$+$equal, and ranked$+$20/80.
Quadratic voting and equal power distribution enhanced participants' trust in the decision-making process, reducing concerns about unexpected outcomes.  As participants shared \textit{``I split my votes across multiple issues, but I think this is the purpose—to vote carefully for the option I care about most. It allows stronger opinions on some issues.  the square thing I like, so even if sometimes someone had more token than me, that's actually not the number that would apply rather square root.''}
A linear regression analysis confirmed this effect: the coefficient for quadratic $0.4772 (P = 0.013)$, for same was $0.4002 (P = 0.038)$. The linear regression considering the interaction also demonstrated statistical significance; the coefficient of \textit{quadratic$\times$same} was $1.1548$ with a P-value of $0.002$.  



\subsection{Quality of Decision-Making Process of Different Democracies.} 
\begin{figure}[h]
    \centering
    \includegraphics[width=\linewidth]{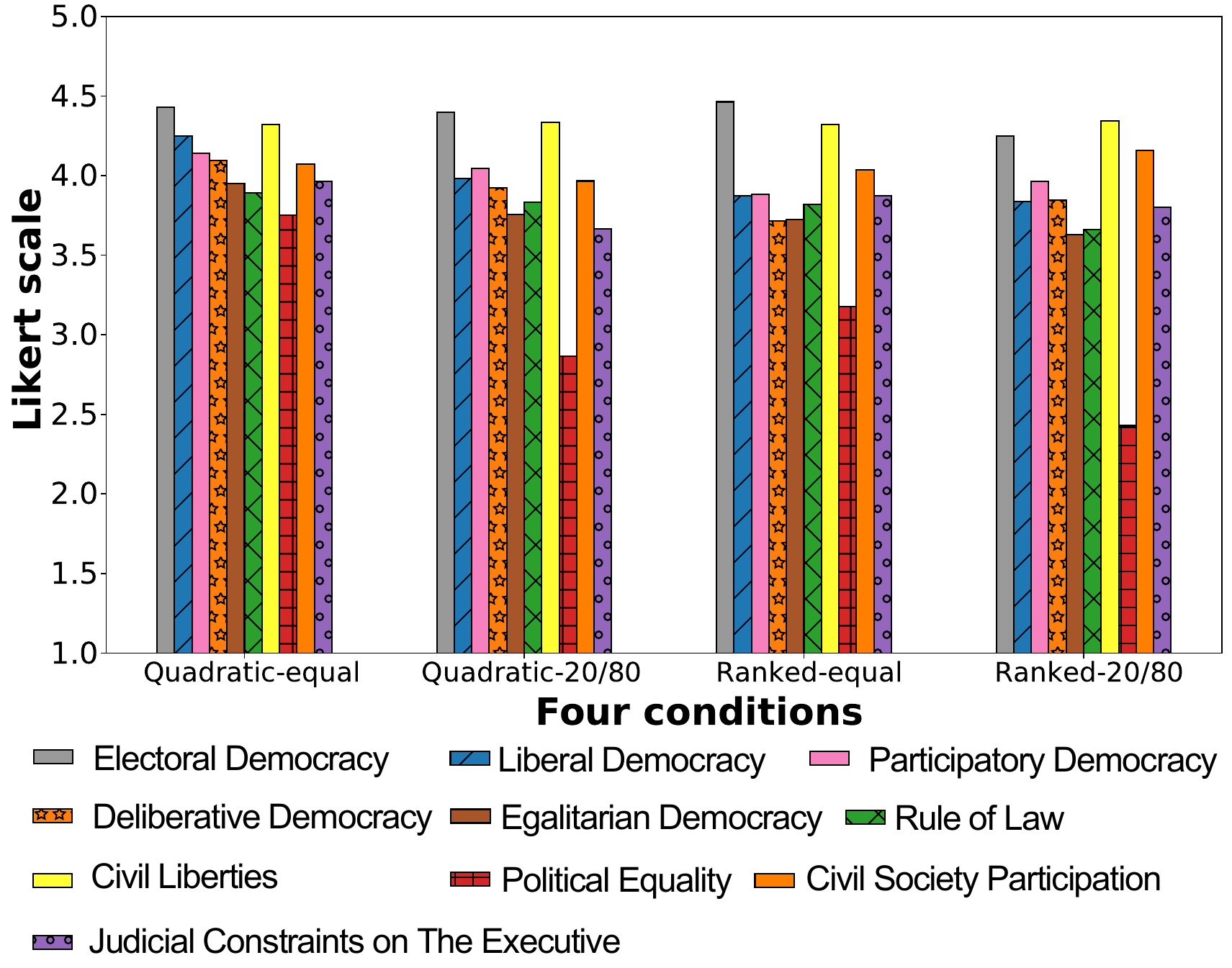}
    \caption{Users' perception of a voting mechanism (obtained through the V-Dem question lists)}
    \label{fig:vdem}
\end{figure}

We examined participants' perceptions of LLM governance using the Varieties of Democracy (V-Dem) (Figure~\ref{fig:vdem}). xf.
As noted, \textit{``The voting was inclusive—I would like this process in chatGPT like system where they broadcast such voting time to time to get some signal from users rather deploying by themselves only.''} This supports the argument that active user participation in AI decision-making can enhance legitimacy, rather than centralized deployment by developers. 
Voting power distribution further reinforced perceptions of political equality (coefficient$=0.8091$, P-value$<0.001$), with linear regression considering interaction confirming its significance (coefficient of \textit{same}$=0.7500$, P-value$=0.019$). This highlights the need of fair representation in AI oversight, where users regardless of their expertise or influence should have a say in shaping AI behavior.

\subsection{Relationship Between Users' Value Towards AI and Their Perceived Democracy Value}
Participants who found LLM personally relevant were more likely to view the DAO-enabled voting process as highly participatory (Figure~\ref{fig:mat})(Pearson Corr$=-0.4426$, P-value$<0.001$). This underscores the need for AI systems to establish personal relevance with users, potentially through more user-centered political content moderation.
Perceptions of deliberative democracy were strongly linked to trust in AI companies(Pearson Corr$=0.4422$, P-value$<0.001$) and perceived AI risks (Pearson Corr$=0.5142$, P-value$<0.001$). This suggests that skepticism about AI risks coexists with the belief that AI governance should involve ongoing public discourse. For LLM governance, this emphasizes the need for mechanisms that allow users to contest, audit, and deliberate on AI-generated political content, rather than simply consuming it.
Participants who valued civil liberties also emphasized the importance of diverse datasets in AI training (Pearson Corr$=0.4646$, P-value$<0.001$), uncertainty handling by AI developers (Pearson Corr$=0.5326$, P-value$<0.001$), the perceived AI risks (Pearson Corr$=0.4407$, P-value$<0.001$), and the desired reliance on AI (Pearson Corr$=0.4950$, P-value$<0.001$). This underscores the necessity of dataset diversity, bias mitigation, and AI uncertainty management in political content generation.

\subsection{Attitude Towards Voting Mechanisms.}
Participants had key attitude in applying voting mechanisms to MM-LLM governance, including (1) progressive and fair process, (2) methods to show the strength of preference, (3) support multiple choices with a unique voice, (4) quantifying perception, (5) Inclusive.AI as practical AI governance applications (e.g., aligning with public preferences). We also found differences in the perception among political parties. Republicans tended to feel significantly more that they could contribute to shaping the space of generative AI models through this process when compared to Democrats (linear coefficient$=0.521,$ P-value$=0.007$). Qualitative analysis of survey data also revealed some differing perspectives (Figure~\ref{tab:politic}). Democrats emphasized empowerment, ease of use, and engagement emphasizing positive experience.  In contrast, republicans prioritized functional and individual priorities like flexible voting power designs, on option to quantify perception through voting, and curiosity. Republican and independent participants also raised concerns about complexity, external influences (majority bias as a good way to go), and post-vote transparency regarding AI developers’ implementation of decisions. 
 However, these findings are not indicative of broader political divisions due to the low frequency of such experiences.

\vspace{-2mm}
\section{Discussion}
\vspace{-1mm}
Our findings underscore two key recommendations for practitioners aligning with LLMs and how to engage users in governance.  

\subsection{DAO as a Technical LLM Governance Solution. }
Transparency in LLM design decisions is utmost importance for aligning AI systems with societal expectations~\cite{mitchell2019model, liesenfeld2023opening}. To do that, it's crucial not only to gain a deeper understanding of public perceptions regarding AI but also to devise methods that actively involve the community in the decision-making processes governing AI technologies. Inclusive.AI tool, underpinned by the DAO mechanism, offers an avenue to actively involve people in governing llm with empirical evidence while presented with a sensitive topic like political video. DAO mechanisms, as digital-first entities, employ mechanisms like initiating proposals, nuanced voting methods, and blockchain-based coordination~\cite{sharma2023unpacking}, offering a structured approach to AI governance~\cite{koster2022human}, a concept endorsed by industry leaders such as OpenAI, Meta, and federal agencies~\cite{openai, biden2023executive}.

A standout feature was our system's Voting method, in which participants found effectively representing their voice directly impacting AI model decisions for future improvements~\cite{arts2004policy}. Participants recognized the potential of these methods in helping developers and government bodies align more closely with public preferences.  However, skepticism remains about whether their votes would translate into real changes in AI models. This highlights the need for government-level guidelines to ensure system compliance and evidence through future audits. 

\subsection{Continuous Human Involvement for LLM Model Adaptability.} 
Our research, drawing on insights from experts in news production reveals that video analysis in media coverage remains largely reliant on manual processes and human intervention. 
Critical frameworks like positionality \cite{callison2019reckoning} and selective exposure balance are essential for ensuring accurate and contextually rich video interpretation, particularly in political reporting. Experts emphasize the need for diverse perspectives and contextual depth to prevent biases and ensure political content reflects a broad spectrum of viewpoints \cite{blumler1999third, jacobs2011space}

Our findings from public deliberation and LLM governance decision-making illustrate how political affiliation shapes perceptions of LLM-generated political content. For instance, Republicans were less likely than Democrats to vote for providing more specific facts and instead favored integrating a user feedback loop for LLM improvement. InclusiveAI platform facilitate on imitating natural human interaction among people acknowledging that conflicting interests and preferences. Rather than seeking consensus on the topic, participants engaged in discussions that helped them identify compromises and make informed voting decisions. 
We suggest that inclusive AI systems could be integrated into LLM tools, such as ChatGPT, allowing users and experts to propose real-time adjustments and engage with broader communities when necessary. It highlights a potential future where people continuously engage in shaping  functionality of AI systems with evolving needs.
Potential risks of this research include political bias reinforcement due to differing perceptions of LLM-generated content, potentially deepening ideological divides. 



\subsection{Future Deployment and Evaluation of Inclusive.AI}

 Stakeholders involved in the governance decision-making in LLM improvement felt their contributions were meaningful, particularly in the configurations of quadratic voting with equal power distribution facilitates minority voices.  These results indicate that a DAO-enabled governance process can increase stakeholder satisfaction in AI decision-making compared to centralized approaches, where only a very small group of stakeholders (e.g., frontier AI companies) make the decision. 
 However, we acknowledge that deployment of our system in the wild can further broaden the impact.  One concrete way that we propose to incorporate Incluisve.AI tool with LLM applications (e.g., ChatGPT, DALL$\cdot$E, Perplexity, etc) or LLM-powered application (e.g, health chatbot, legal interpretation chatbot) is that Inclusive.AI could be integrated as a plugin within ChatGPT that enables users to challenge and collaboratively deliberate over AI-generated content whether it is a summary of a political debate, a medical explanation, an image interpretation, or a cultural analysis. When users find an output unsatisfactory, biased, or harmful, they can press a \textit{``Challenge this Response''} button, opening a structured interface to provide alternative views, engage in a real-time deliberation thread, and vote on LLM improvement suggestions. The system can capture people's varying perceptions 
and flags contested outputs for model refinement. Over time, it can aggregate user preferences across domains, politics, health, education, and social issues to inform transparent and participatory AI behavior tuning.

To extend Inclusive.AI beyond this study, we envision a set of realistic studies modeling contentious AI outputs that might provoke public disagreement. In one scenario, an AI system generates misleading vaccine safety information~\cite{ahmad2025vaxguard,xu2025mmdt}--- Inclusive.AI can enable deliberation among patients, public health experts, and ethicists, evaluating outcomes such as perceived legitimacy while media outlets can integrate it to audit neutrality in political captioning~\cite{edenberg2023disambiguating,norris2009public}; and civic platforms may apply it to generative summaries of legislation~\cite{fan2020digital, tsai2024generative}. In the future, we aim to evaluate Inclusive.AI’s deliberative refinement model against top-down moderation~\cite{seering2020reconsidering}, non-deliberative crowd voting~\cite{fishkin2003quest, gerber2014deliberative, sharma2022s}, and expert-only audits~\cite{zhang2020us, costanza2022audits} to assess trust, transparency, and correction efficacy.

\subsection{Ethical Considerations in Democratic AI Governance}
Inclusive.AI in practice requires careful attention to ethical and legal factors to ensure that the governance process itself upholds inclusivity, fairness, and complies with regulations while people's identity and interaction are involved. Our system addresses these challenges by minimizing personal data on-chain and using pseudonymized identifiers for participants to engage in deliberation and democratic voting in LLM improvement. Depending on the use case, one can enable a setting to keep sensitive user information (e.g., demographic details or discussion content) off-chain, while decision records remain immutable and transparent. Additionally, to further ensure privacy, current infrastructure can adapt with emerging solutions like confidential DAOs which encrypt governance data yet maintain transparent, immutable logs~\cite{zamaConfidentialGovernance, sharma2024can} as well as maintain confidence among stakeholders that participation will not expose their personal data and will comply with privacy regulations (e.g., GDPR ~\cite{haque2021gdpr,potter2025gap}) and ethical norms.

Another critical consideration is preventing the concentration of power and ensuring fairness within the DAO itself. A known pitfall in some blockchain governance systems is the risk of “whale” voters or token-based plutocracy, where a small number of actors accumulate disproportionate influence~\cite{sharma2023unpacking, brookingsHiddenDanger}. We mitigate this risk through system design, such as, quadratic voting to curb dominance by any single group. For example, quadratic voting is used to give diminishing returns on additional votes, so that participants must spend votes quadratically to express stronger preferences~\cite{lalley2018quadratic}. This mechanism ensures that minority opinions can still influence outcomes on issues they care deeply about, counterbalancing majority rule. In practice, quadratic voting allowed marginalized voices in our study to impact AI design decisions significantly more than under a traditional one-person-one-vote system. In the future, we will explore alternatives to purely token-based governance or a hybrid approach, for instance, sortition-based DAOs which enable a rotating panel of randomly chosen participants, as a \textit{``governance jury,''} for unbiased participant selection, ensuring no clique can continuously dominate decisions~\cite{kelsey2023implementing}. These combined measures are compatible with the current infrastructure of Inclusive.AI to prevent power concentration with fair allocation of voice, advanced voting schemes, and hybrid governance models. 

Finally, we designed the system to embed values with traceability throughout the AI governance lifecycle. It addresses the ``black box'' concern often raised in AI ethics since decisions made via Incluisve.AI are traceable in blockchain.  Stakeholders and auditors can inspect how a particular AI model update was agreed upon, which arguments were presented. This not only builds trust but also facilitates external oversight or audits for compliance. In case of regulatory review or legal disputes, the immutable governance record serves as evidence of due diligence and community consent in the AI’s decision-making process.

\section{Conclusion}

Our study demonstrates that decentralized, participatory governance mechanisms in Inclusive.AI platform can support large language model (LLM) development decision making with diverse public values, particularly in politically sensitive contexts. By combining individual and collective deliberation with structured voting methods like quadratic and weighted voting, the platform enabled meaningful engagement from users across political affiliations. 
Our research suggests that integrating democratic input into AI governance not only enhances user satisfaction but also offers a scalable model for aligning AI behavior with societal expectations.

\section{Ethics Statement}
This study protocol involving human  subjects was approved by the Institutional Review Board (IRB). The data collection and transcription generation was anonymous to preserve privacy of the users. This study explores decentralized governance mechanisms in decision-making for LLM improvement by engaging users, particularly in politically sensitive contexts. The InclusiveAI tool with transparent design, equitable participation can allow to shape AI with broader perspectives. This also has a future potential to potentially involve regulatory oversight for for the responsible implementation.






\section{Positionality Statement}
All authors are currently affiliated with US academic institutions, and all our interviewees also live in the United States. Although none of the team members have extensive experience in politics or journalism, the research team includes members who have long been engaged in AI/ML, Human Centered Computing and privacy and security research, as well as members with extensive experience in communication and decentralized applications research experience, which ensures that we can understand the problems faced in AI governance from multiple perspectives.

\section{Adverse Impact Statements}
As the paper only contains non-identifiable interview, survey, deliberations and voting tally. All of the data is collected with anonymity, which is one of the fundamental functionalities of the decentralized system infrastructure implemented in Inclusive.AI. We do not expect that the dissemination of this paper will have a substantial adverse impact. 

\bibliography{aaai25}
\appendix


  \section{Appendix}
  \label{pilot}
\textbf{Pilot Experiment. }We conducted a pilot study using the "Inclusive.AI" tool to facilitate deliberation on AI-related value topics presented in video format. For this pilot, the topic was political, using a clip from the 2020 U.S. presidential debate. The tool guided users through a process of eliciting their values and expectations for LLM outputs on sensitive topics via individual and group deliberation. It also provided a platform for users to share their preferences on how MM-LLMs should function in the future, as detailed in Section~\ref{design-deliberation}.

The pilot study provided valuable insights to improve the tool. For example, the group deliberation feature initially didn't include suggested topics, which led to difficulties in starting discussions. Participants recommended including suggested topics in the group-live chat feature, leading us to a design update in the Inclusive.AI tool for the main experiment.
The pilot also helped refine the MM-LLM update options used to gather participants' preferences in democratic decision-making. The initial options, based on the literature, included:
(a) Use the current model as is,
(b)Context-Aware Adaptation,
(c)Use Feedback Loop Integration, and
(d)Advanced Modality Integration Technique.
Feedback from the pilot deliberation on the political video topic led to revisions for the main study, resulting in the following refined options:
(a)Use the current model as is,
(b)Provide more specific facts,"
(c)Integrate user feedback loop," and
(d)Provide analysis of the speaker's emotion and sentiment.
We performed a thematic analysis of the pilot data to identify key themes in the deliberations.

\textbf{Selection of Video Content.}
To select a video sample of political content for this experiment on user expectations of MM-LLM interpretation, we applied specific criteria to ensure relevance.

First, we considered the \textit{relevance to current events} within the field of communication. This means that the content should be timely and relate to significant political events, such as election cycles, with the 2024 presidential election being a major focus.

Second, we included a diversity of perspectives spanning a broad range of political views. This meant selecting videos that included different ideologies or interest groups to obtain a holistic view of the discourse

Third, we looked at the contextual depth or complexity of the visuals in the video. Videos should not only present political events or opinions, but also provide context that helps in understanding background information, historical ties, and potential implications. We also considered the diversity of video visuals and interactions to ensure that they provide sufficient context for MM-LLMs to generate interesting interpretations.

Fourth, we ensure the authenticity and reliability of the content sources by considering authoritative channels such as the top three cable news channels in the US (CNN, FOX, and MSNBC)and popular datasets of political videos used in the existing literature. We also considered the timeline of videos from 2016 to 2024 considering two major election cycles and major technological advancements in AI and digital media. 
Finally, to safeguard against misleading or harmful content, particularly in sensitive areas like politics, we manually reviewed the video content.

\textbf{How Experts Would Prompt to Analyze the Video?}
Experts suggested various ways they would prompt ChatGPT to analyze a presidential debate. Most would start with a general question like, \textit{Can you help me to summarize what they are talking about?''} E1 mentioned that she would first ask for a summary and then follow up with, \textit{``If I get the output, I might ask something else.''} Similarly, E2 would prompt, \textit{``Provide a summary of the videos and the point of each person in this content.''}
Two experts, like E3, preferred more detailed instructions, saying, \textit{`Give me a short news brief about the presidential debate between Trump and Biden about health care policy and Obamacare. Also, capture some of the tough visual aspects, describe some of the back-and-forth banter between the moderator and Trump, and the personal attacks where people can't get a word in.''} E6 would ask meta-questions to utilize ChatGPT in a journalism context: \textit{``I normally wouldn't use ChatGPT to analyze only one video unless I have a hundred. I would want to see patterns across videos. I would ask it to analyze how many times there are interruptions, how long candidates talk over each other, and other specific metrics.''}

\section{Experts’ Interview}
\label{experts}

In this section, we present the questions that we asked during the expert interview.


\subsection{General Introduction}
\begin{enumerate}
    \item Could you briefly talk about your primary area of expertise in communication, journalism, or media studies?
    \item What kind of media do you usually work on? Do you ever work on video content? Can you share a recent experience with video content and describe what it was about?
    \item How do you choose videos for your work? (This question is based on an earlier response about the type of work they do with video.)
\end{enumerate}

\subsection{Assessment of Videos by Communication, Media Scholars, or Journalists}
\begin{enumerate}
    \item When reviewing events (e.g., live or video) related to complex subjects such as politics, what key factors do you consider in drafting/generating an article on this event?
    \item What does "good video analysis" mean to you?
    \item Can you walk me through the process you employ to analyze video content to write a report?
    \item Please consider this video (a provided video). Feel free to analyze it manually or with any existing tool you typically use.
    \item Could you please write your interpretation of this video content and share it afterward?
\end{enumerate}

\subsection{Perceptions of Large Language Models (LLMs) for Video Analysis and Assessment of LLM-Generated Video Analysis}
\begin{enumerate}
    \item How familiar are you with the use of large language models (LLMs), such as ChatGPT, for video analysis?
    \item What do you think about the idea of using LLMs for video analysis? What are the pros and cons in your opinion?
    \item \textbf{Demo:} Show a sample video analysis generated by ChatGPT.
    \item Now read this analysis result from the LLM for the same video. What do you think about this analysis?
    \item If the analysis did not match your expectations, why?
    \item If the analysis met your expectations, why?
    \item Based on your review of the video and the LLM-generated response, what criteria do you consider necessary for a good video analysis result?
    \item If you were to use ChatGPT for video analysis, how would you prompt it?
\end{enumerate}

\subsection{Use Cases of AI in Communication and Journalism}
\begin{enumerate}
    \item Can you share any current use cases where AI has been effectively integrated into communication or journalism practices?
    \item How do you see the role of AI evolving in the field of journalism and media studies over the next five years?
\end{enumerate}

\section{Survey Study Protocol}

In this section, we present the survey questions used in the Inclusive.AI study, which involved 114 participants.

\textbf{Governance Survey Questions}
We'd like to understand your voting experience. Below, we'll present a series of statements related to your voting experience and different voting methods and voting power you have used. Please indicate your level of agreement using the Likert scale provided.

Please use the following scale: 1 = Strongly Disagree, 2 = Disagree, 3 = Neutral, 4 = Agree, 5 = Strongly Agree.
\begin{itemize}
    \item The decision-making process was indecisive.
    \item The decision-making process was good at maintaining order.
    \item The decision-making process may have problems, but it's better than any other form of government.
\end{itemize}

Please rate your attitude toward governance components such as voting methods (e.g., quadratic, ranking), voting power, etc.
\begin{itemize}
    \item I found the voting method (Weighted ranking/Quadratic) meaningful to include my voice.
    \item I felt that I could contribute to shaping the space of the Generative AI model.
    \item I found this voting method relevant to the purpose of the proposal or proposal type.
    \item I found this voting power/token distribution (e.g., equal power, variable power) meaningful in including my voice.
    \item I found the voting method (Weighted ranking/Quadratic) fair.
    \item I felt I have some power to affect change in Generative AI future development.
    \item I found voting power distribution among users equitable.
    \item I felt the voting power distribution can result in unexpected outcomes.
\end{itemize}

Open-Ended Questions
\begin{itemize}
    \item Please explain how you found the voting process to share your preferences on the video analysis by ChatGPT.
    \item What do you think is the impact of your contributions on designing a Generative AI Model that reflects informed public consensus?
    \item What are the potential benefits of personalizing generated video analysis by ChatGPT to align with your preferences?
    \item What are your concerns, if any, about analyzing video by ChatGPT?
\end{itemize}

\textbf{Democratic Decision-Making Measures}

\textbf{Electoral Democracy:}
\begin{itemize}
    \item I believe that the voting process was free and fair.
    \item I felt all users had the right to vote.
\end{itemize}

\textbf{Liberal Democracy:}
\begin{itemize}
    \item I believe AI models will operate independently without interference from the development team.
    \item I felt free to provide feedback on the AI model update without fear of repercussions.
\end{itemize}

\textbf{Participatory Democracy:}
\begin{itemize}
    \item I felt that I had ample opportunities to influence the AI model update process beyond just voting.
    \item I felt that my feedback matters in the decisions made for AI model updates.
    \item I believe AI model update decisions will reflect the needs and preferences of the user community.
\end{itemize}

\textbf{Deliberative Democracy:}
\begin{itemize}
    \item AI model update decisions are made after thorough discussion with the user community.
    \item There is a culture of open dialogue and discussion in the AI model update community.
    \item I believe developers of this AI model will prioritize user interests over their own preferences.
\end{itemize}

\textbf{Egalitarian Democracy:}
\begin{itemize}
    \item I felt, regardless of my background, I have equal influence in the AI model update decision process.
    \item I felt large corporations or specific user groups do not have undue influence over AI model update decision processes.
\end{itemize}

\textbf{Rule of Law:}
\begin{itemize}
    \item I believe developers will be held accountable for flaws or biases in the AI model updates after this decision process.
    \item The decision process treats every user's input equally, regardless of their status.
\end{itemize}

\textbf{Civil Liberties:}
\begin{itemize}
    \item I felt free to express my opinions on AI model updates without fear.
    \item I felt free to participate in any community or forum discussing the AI model update decision process.
\end{itemize}

\textbf{Political Equality:}
\begin{itemize}
    \item Wealthy individuals do not have more political influence than ordinary citizens.
    \item All ethnic and religious groups have equal political rights and influence.
\end{itemize}

\textbf{Civil Society Participation:}
\begin{itemize}
    \item User communities play an active role in shaping AI model update policies.
    \item The development team actively seeks input from user groups and communities.
\end{itemize}

\textbf{Political Ideology:}
\begin{itemize}
    \item What are the three political issues that matter to you?
    \item On a scale of 1 (strongly disagree) to 5 (strongly agree), please rate your satisfaction with the current political climate.
    \item How important is politics in your daily life? (Options: Very important, Somewhat important, Neutral, Not very important, Not at all important)
    \item Which political party do you most identify with? (Options: Republican, Democratic, Independent, Libertarian, Green, Other)
    \item How would you describe your political orientation? (Options: Very conservative, Somewhat conservative, Moderate, Somewhat liberal, Very liberal, Not sure, Prefer not to say)
\end{itemize}

\textbf{Demographic Questions:}
\begin{itemize}
    \item What is your age range?
    \item What is your gender identity?
    \item Are you currently enrolled in an educational institution?
    \item What is your highest level of education completed?
    \item Please select your racial or ethnic background.
    \item How frequently do you use technology or digital devices?
    \item How often do you use an AI assistant such as ChatGPT?
\end{itemize}

\textbf{AI Value Questions}

\textbf{Likert Scale on AI Representation and Customization:}
\begin{itemize}
    \item AI models should prioritize generating diverse outputs to represent a wide range of individuals.
    \item Customization options, like specifying gender or ethnicity, are vital for inclusive AI-generated videos.
    \item A diverse dataset in AI training is essential to prevent bias and ensure fair representation when analyzing videos on political topics.
    \item AI developers should prioritize uncertainty handling to avoid assumptions and ensure diverse outputs.
\end{itemize}

\textbf{Trust and Personalization of Generative AI:}
\begin{itemize}
    \item The use case is not relevant to me.
    \item I feel AI could infringe on my representation.
    \item I do not fully trust the abilities of an AI model.
    \item The use case is too important to let the AI model decide for me.
    \item I am concerned that it would not be exactly clear how video analyses are produced by AI.
    \item I believe that AI in general would treat me fairly when making decisions and suggestions.
    \item If I have any problem with AI decisions, I believe OpenAI would take necessary measures.
\end{itemize}

\onecolumn  
 \subsection{Figures and Tables}
\begin{figure*}[!htb]
\begin{subfigure}{0.50\textwidth}
  \includegraphics[width=\linewidth]{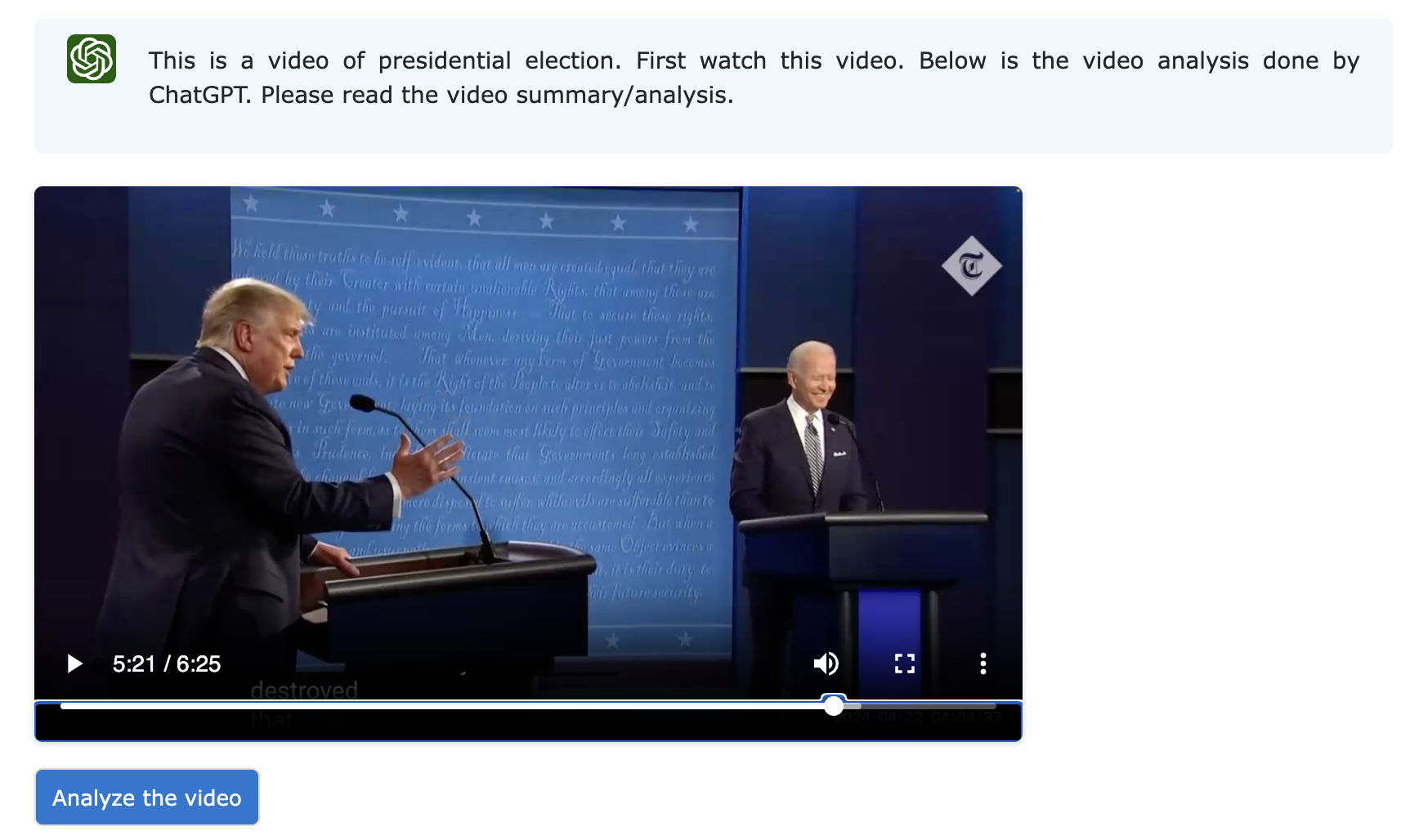}
  \caption{AI value topic in video format where value topic is political- a clip from US presidential debate 2020}
\end{subfigure}\hfill
\begin{subfigure}{0.50\textwidth}
  \includegraphics[width=\linewidth]{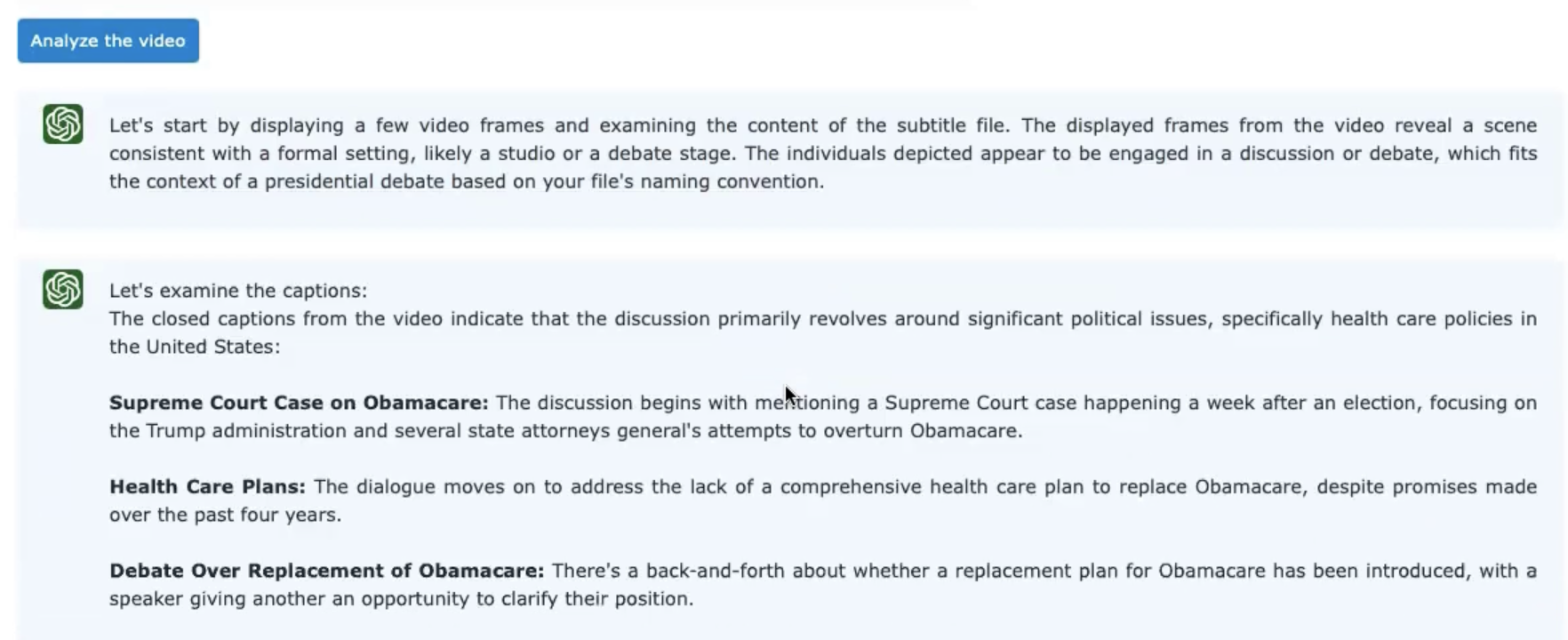}
  \caption{Generated response of ChatGPT based on the video with multiple data (e.g. video frames, audio, closed caption) }
\end{subfigure}
\caption{Users' workflow in Human-AI interaction on AI value topic}  
\label{user-workflow}
\end{figure*}
\begin{figure*}[h]
    \centering
    \includegraphics[width=0.9\linewidth]{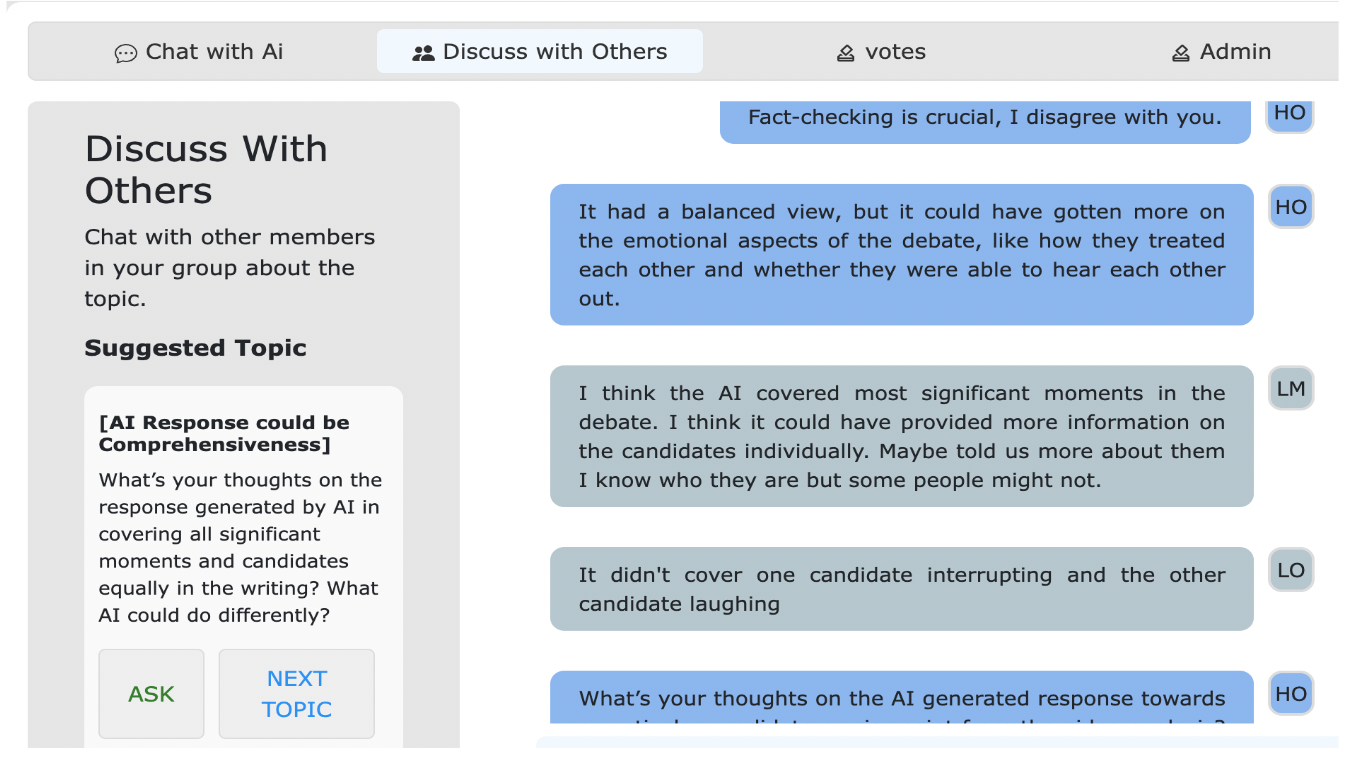}
    \caption{Discussion interface} 
    \label{fig:discussion}
\end{figure*}

\begin{figure*}[h]
    \centering
    \includegraphics[width=\linewidth]{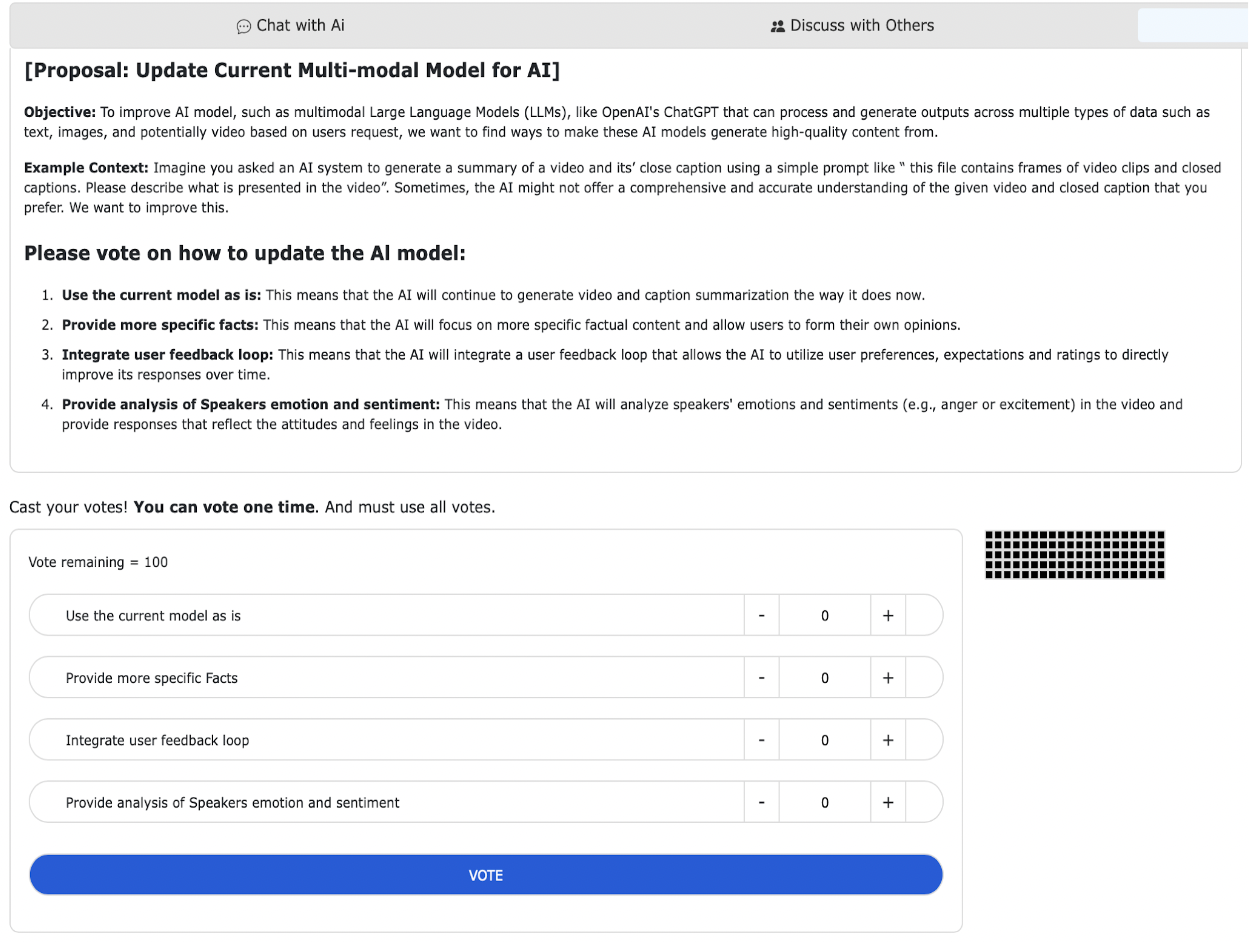}
    \caption{Governance decision making voting interface: Treatment condition example of quadratic voting interface for MM-llm decision for future improvement} 
    \label{fig:vote}
\end{figure*}

\begin{figure*}[h]
    \centering
    \includegraphics[width=1.0\linewidth]{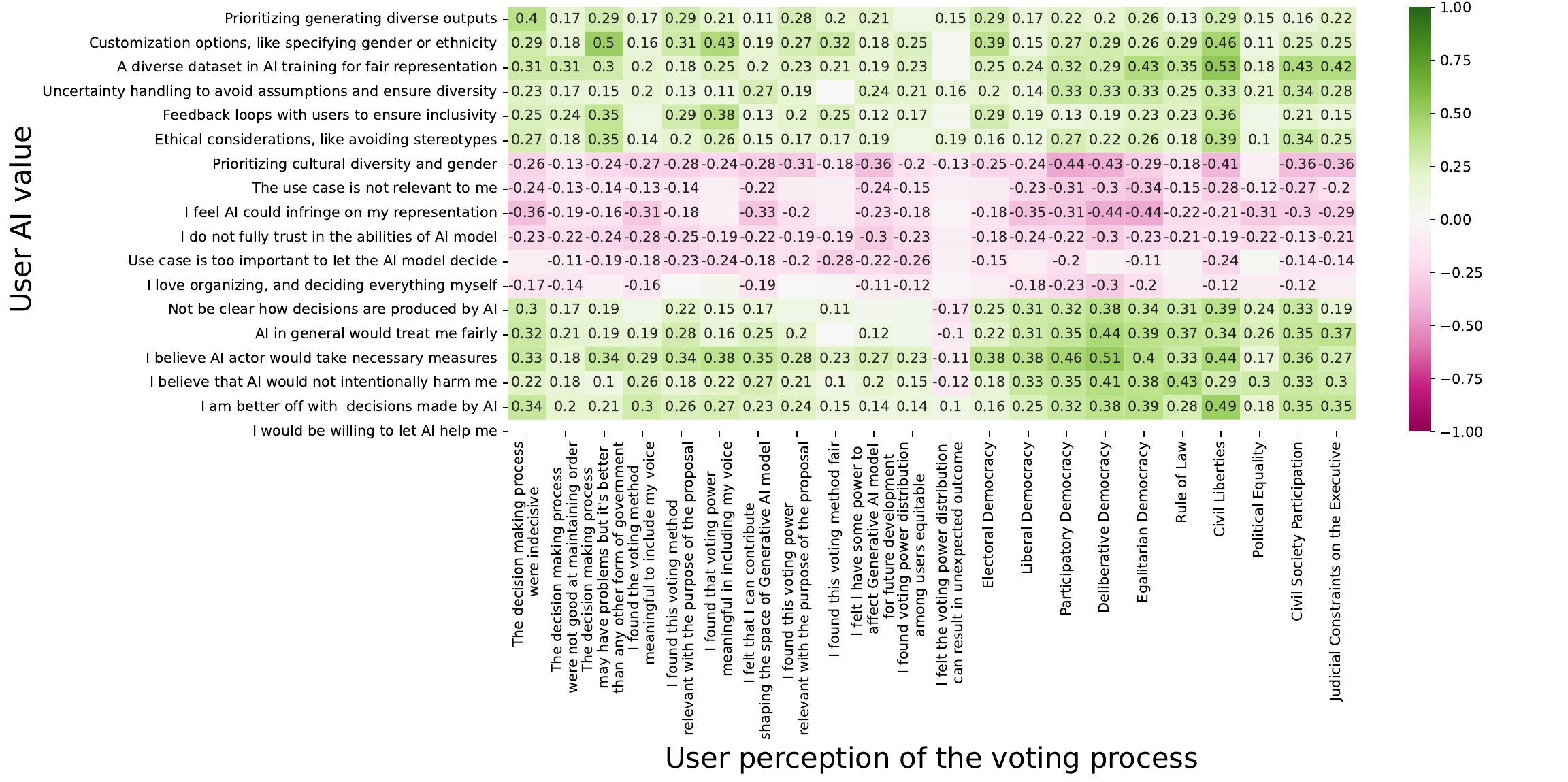}
    \caption{Correlation matrix of users' perceived quality of democracy (V-Dem Likert scale) with the predictor's variables that are users' perceived values on AI topics (Likert scale) including constructs, such as trust, perceived fairness, perceived accountability, and expected personalization.}
    \label{fig:mat}
\end{figure*}

\end{document}